\documentclass[pdflatex,sn-mathphys-num]{sn-jnl}

\usepackage{graphicx}%
\usepackage{multirow}%
\usepackage{amsmath,amssymb,amsfonts}%
\usepackage{amsthm}%
\usepackage{mathrsfs}%
\usepackage[title]{appendix}%
\usepackage{xcolor}%
\usepackage{textcomp}%
\usepackage{manyfoot}%
\usepackage{booktabs}%
\usepackage{array}
\usepackage{algorithm}%
\usepackage{algorithmicx}%
\usepackage{algpseudocode}%
\usepackage{listings}%

\theoremstyle{thmstyleone}%
\theoremstyle{thmstyletwo}%

\theoremstyle{thmstylethree}%

\begin{document}

\title[Article Title]{MvBody: Multi-View-Based Hybrid Transformer Using Optical 3D Body Scan for Explainable Cesarean Section Prediction}

\author*[1]{Ruting Cheng}\email{rcheng77@gwu.edu}

\author[1]{Boyuan Feng}\email{fby@gwu.edu}

\author[1]{Yijiang Zheng}\email{yijiangzheng@gwu.edu}

\author[1]{Chuhui Qiu}\email{chqiu@gwu.edu}

\author[1]{Aizierjiang Aiersilan}\email{alexandera@gwu.edu}

\author[2]{Joaquin A. Calderon}\email{joacocal91@gmail.com}

\author[3]{Wentao Zhao}\email{zhaowentao@toki.waseda.jp}

\author[4]{Qing Pan}\email{qpan@gwu.edu}

\author[1]{James K. Hahn}\email{hahn@gwu.edu}

\affil*[1]{\orgdiv{Department of Computer Science}, \orgname{The George Washington University}, \orgaddress{\street{800 22nd Street NW}, \city{Washington DC}, \postcode{20052}, \state{}, \country{USA}}}

\affil[2]{\orgdiv{Department of Obstetrics and Gynecology}, \orgname{The George Washington University}, \orgaddress{\street{2150 Pennsylvania Ave. NW}, \city{Washington DC}, \postcode{20037}, \state{}, \country{USA}}}

\affil[3]{\orgdiv{Department of Integrative Bioscience and Biomedical Engineering}, \orgname{Waseda University}, \orgaddress{\street{Takadanobaba, Shinjuku-ku}, \city{Tokyo}, \postcode{162-0075}, \state{}, \country{Japan}}}

\affil[4]{\orgdiv{Department of Biostatistics and Bioinformatics}, \orgname{The George Washington University}, \orgaddress{\street{800 22nd Street NW}, \city{Washington DC}, \postcode{20052}, \state{}, \country{USA}}}

\abstract{Accurately assessing the risk of cesarean section (CS) delivery is critical, especially in settings with limited medical resources, where access to healthcare is often restricted. Early and reliable risk prediction allows better-informed prenatal care decisions and can improve maternal and neonatal outcomes. However, most existing predictive models are tailored for in-hospital use during labor and rely on parameters that are often unavailable in resource-limited or home-based settings. In this study, we conduct a pilot investigation to examine the feasibility of using 3D body shape for CS risk assessment for future applications with more affordable general devices. We propose a novel multi-view-based Transformer network, MvBody, which predicts CS risk using only self-reported medical data and 3D optical body scans obtained between the 31st and 38th weeks of gestation. To enhance training efficiency and model generalizability in data-scarce environments, we incorporate a metric learning loss into the network. Compared to widely used machine learning models and the latest advanced 3D analysis methods, our method demonstrates superior performance, achieving an accuracy of 84.62\% and an Area Under the Receiver Operating Characteristic Curve (AUC-ROC) of 0.724 on the independent test set. To improve transparency and trust in the model’s predictions, we apply the Integrated Gradients algorithm to provide theoretically grounded explanations of the model’s decision-making process. Our results indicate that pre-pregnancy weight, maternal age, obstetric history, previous CS history, and body shape, particularly around the head and shoulders, are key contributors to CS risk prediction.
}


\maketitle

\section{Introduction}
\label{sec:intro}

The global rate of cesarean section (CS) deliveries has been rising for decades, with marked disparities across countries and regions \cite{betran2021trends, boatin2018within}. While CS is often overused in high-resource settings without medical necessity due to fear of childbirth, in resource-limited areas, barriers such as restricted access to medical care, inadequate prenatal education, and cultural preferences for home births can delay or prevent medically indicated CS. These delays contribute to preventable maternal and neonatal morbidity and mortality \cite{begum2021global,chien2021global,qadeer2024current, boatin2018within,irwinda2021maternal}. To improve outcomes in underserved populations and to identify accessible risk factors that can inform prenatal care planning, there is a growing need for an explainable and easily deployable predictive model for CS risk, particularly one suitable for community-based or home-based healthcare environments.

Several predictive models have been proposed to assist with CS-related decision-making \cite{grobman2007development, campillo2018predictive, guan2020prediction, lopez2022risk}. However, most of these models are designed for well-equipped medical settings, where clinical biomarkers such as placental or amniotic fluid status are available \cite{campillo2018predictive, guan2020prediction}. Moreover, many models are tailored to specific subgroups, such as women with prior CS or those undergoing labor induction. While those approaches offer valuable insights, their generalizability is limited, and they are not well suited for application in resource-limited settings.

To bridge this gap, we explored the potential of 3D body shape as an alternative data source. Unlike costly imaging techniques or invasive diagnostic procedures, 3D body shape can be obtained using affordable commercial scanners or even mobile applications with 3D reconstruction \cite{ng2016clinical,samavati2023deep}. Prior studies have demonstrated associations between anthropometric features, such as pelvic dimensions, and the risk of CS delivery \cite{boucher2022maternal, alijahan2014diagnostic, korhonen2014diagnostic}. Some researchers have investigated the use of 3D body scans in obstetrics, however, most applications remain focused on visualization or the extraction of traditional anthropometric measurements \cite{gradl2022application,glinkowski2016posture,dathan2023novel}. A recent study took a step further by using 3D scans to predict pregnancy complications and assess maternal and fetal health status, however, it relied on manual pre-processing and low-dimensional 3D shape representation of level circumferences \cite{cheng2025maternal}. Using 3D body shapes captured between 18 and 24 gestational weeks, this method demonstrated limited effectiveness for CS risk prediction, as the scans were taken too early to capture the late-pregnancy body shape changes most relevant to delivery outcomes.

To better leverage the predictive potential of 3D body scans, we investigated various 3D processing techniques, including voxel-based, point-cloud-based and multi-view-based methods \cite{zhao2024point, wang2024afrnet, zeng2021hierarchical, su2015multi, chen2021mvt, wang2022ovpt}. These methods have demonstrated strong performance on relatively simple synthetic datasets. However, practical medical applications involving detailed human body representations remain underexplored. In this context, multi-view-based approaches offer notable advantages because they can leverage mature 2D image processing architectures and benefit from pretraining on large-scale, real-world image datasets. This design improves training efficiency and enhances performance when working with small, task-specific medical datasets. Meanwhile, metric learning has shown superior efficacy in data-scarce scenarios \cite{oreshkin2018tadam, jung2022few, li2023deep}.  Therefore, we propose an innovative multi-view-based hybrid framework incorporating metric learning for predicting CS risk using 3D body scans and self-reported medical data. Our main contributions are summarized as follows:

\begin{itemize}
\item We propose MvBody, a neural network that combines the Convolutional Neural Network (CNN) and Transformer blocks to predict the risk of CS. MvBody utilizes multi-view projections of 3D body shapes alongside self-reported medical data, providing a complementary tool to existing clinical risk assessment strategies.
\item To address the challenge of limited sample size, which is common in medical deep learning applications, we incorporate pretraining and metric learning to improve the model performance and generalizability.
\item We provide detailed explanations for the decision making using Integrated Gradients (IG) algorithm, enabling clear identification of risk factors and deeper insight into the network’s reasoning. Our visualization approach highlights contributing features at both the element level (medical features) and the pixel level (image features) to illustrate which medical parameters and body regions are most relevant to the prediction.
\item To the best of our knowledge, this is the first study to employ 3D body scan data and a deep neural network to predict CS risk in the general population without relying on professional medical equipment.

\end{itemize}

\section{Related Work}
\label{sec:relatedwork}

Maternal anthropometric measurements have long been associated with CS risk. Solomon et al. identified that the risk of Cephalopelvic Disproportion (CPD)-related CS was associated with maternal height, foot length, Michaelis horizontal diagonal, and head circumference values \cite{solomon2018age}. Similarly Boucher et al. found correlations between CS risk and variables such as weight, Body Mass Index (BMI), waist circumference and biceps skinfold thickness \cite{boucher2022maternal}. Campillo-Artero et al. evaluated both Logistic Regression (LR) and Random Forest (RF) models to predict emergency CS, incorporating variables like intrapartum pH as risk factors \cite{campillo2018predictive}. Recently, 3D optical body scans have been introduced into obstetric research. Gleason et al. extracted a set of anthropometric measurements from 3D body scans to assess the CPD-related obstructed labor risk and found their results comparable to both Magnetic Resonance Imaging (MRI)-based measurements and traditional anthropometry \cite{gleason2018safe}. Dathan-Stumpf et al. validated the utility of 3D body scan–derived measurements for evaluating vaginal breech delivery feasibility, comparing their results with pelvimeter and MRI pelvimetry data \cite{dathan2023novel}. Despite these pioneering studies explored the usability of 3D body scanning in obstetrics, innovative use of this new data modality and corresponding advanced algorithms has yet to be fully studied. 

However, successful applications of 3D body scanning in other medical domains, such as nutritional assessment, have highlighted its potential to extract clinically relevant information from body shape. Lu et al. represented 3D body shapes using level circumferences and applied a Bayesian network to predict pixel-level body composition and body fat percentage on the 2D projections of 3D scans \cite{lu20193d}. Wang et al. constructed a “Shape Map” descriptor using Computed Tomography (CT)-derived body contours to predict total and visceral fat percentages \cite{wang2021pixel}.  Feng et al. designed a part-to-global Multilayer Perceptron (MLP) that used appendicular composition estimates to improve body composition estimation, while Zheng et al. applied a Transformer-based network to process raw 3D point clouds of body scans for fat percentage assessment\cite{fengenhanced, zheng2024d3bt}. These studies offer valuable methodological insights for applying advanced representations and algorithms to 3D shape analysis in obstetrics. A recent study by Cheng et al. explored more innovative algorithms and 3D shape representations in obstetrics field \cite{cheng2025maternal}. Using a hybrid recurrent neural network (RNN), the authors analyzed dense abdominal level circumferences extracted from 3D body scans to estimate fetal weight and predict pregnancy complications. However, their approach was limited by its low-dimensional shape representation, which led to the inevitable loss of critical body structural details.

To better preserve body shape details, more comprehensive 3D representation should be considered. According to the representation formats of 3D data, existing algorithms can be mainly divided into three categories: point-based methods, voxel-based methods, and projection-based methods \cite{guo2020deep, qi2017pointnet, wu20153d, su2015multi}. Point-based methods directly learn features from raw 3D point clouds and show high potential, but they still rely on carefully designed feature extractors \cite{qi2017pointnet, qi2017pointnet++}. Voxel-based methods leverage established algorithm architectures originally developed for pixel data. However, increasing voxel resolution leads to substantial memory costs and risks of overfitting, making them less suitable for small datasets \cite{wu20153d, zhang2022pvt}. Projection-based methods project 3D data into a series of 2D images, allowing the use of mature neural networks designed for 2D vision tasks \cite{su2015multi, zeng2021hierarchical}. This approach enables the reuse of well-established architectures such as VGG, along with the pretrained parameters from large-scale 2D dataset \cite{simonyan2014very}. It is a big advantage for applying deep learning model in medical scenario with limited data. With the developments of 2D image processing networks, the projection-based methods are also evolving. Zeng et al. introduced a cross-channel feature aggregation module to improve learning efficiency \cite{zeng2021hierarchical}. Chen et al. encoded image patches into token sequence and used the Transformer to capture contextual information \cite{chen2021mvt}. Wang et al. extended this concept by incorporating an entropy-based optimal viewset construction layer to further enhance performance \cite{wang2022ovpt}. Although these studies primarily focused on improving computational efficiency on synthetic datasets rather than capturing subtle shape details on real-world datasets, they inspired the design of our current model architecture. Considering the multi-modal nature of our dataset, we also bridged the processing streams of 3D body scans and self-reported medical information to predict CS risk. Moreover, we incorporated metric learning to capture subtle differences between body shapes in the feature space, which further enhanced model performance under imbalanced and small-sample conditions.

\section{Dataset}
\label{sec:dataset}

In our study, we recruited 101 participants based on the following exclusion criteria: (1) age under 18; (2) multiple gestations; (3) diagnosis of an enlarged fibroid uterus; (4) BMI greater than 60; (5) presence of any unstable medical or emotional condition or chronic disease that could interfere study participation; (6) history of body altering procedures such as liposuction or plastic surgery. The study protocol was approved by the Institutional Review Board (NCR224227, accepted on 08/07/2023), and informed consent was obtained from all participants. All procedures were performed in compliance with relevant laws and institutional guidelines.

Each participant underwent 3D optical body scanning between gestational weeks 31 and 38 using the Fit3D scanner (Fit3D, San Francisco, CA), a device previously validated for high precision \cite{ng2016clinical,sobhiyeh2021digital}. Participants were instructed to wear form-fitting clothing and tightly secure their hair during the scan to ensure accuracy. Each participant was scanned two to three times to improve data reliability, resulting in a total of 236 scans from the 101 participants. In addition to 3D scan data, we collected a set of self-reported and easily measurable clinical variables. These included demographic and obstetric data such as race, age, height, pre-pregnancy weight, gravidity, parity, prior CS delivery, and history of gestational hypertension, gestational diabetes mellitus, and pre-eclampsia. We also recorded the presence of chronic pre-existing conditions (e.g., asthma, hypertension, diabetes mellitus), as well as current weight and gestational age at the time of the scan. The type of delivery served as the prediction target, and all CS deliveries were medically indicated. Among the 101 participants, 25 delivered via CS, and 76 had vaginal deliveries. To facilitate comparison with traditional machine learning approaches, we also extracted anthropometric measurements automatically generated by the Fit3D scanner. These included circumferences of chest, waist, hips, biceps, forearms, thighs, calves, trunk-to-leg volume ratio, waist-to-hip ratio, and a body shape rating value. These features, combined with the same set of clinical variables, were used as inputs for the baseline machine learning models evaluated in our study.

\section{Methods}
\label{sec:methods}

In this section, we present our novel network architecture, the loss function for metric learning, and the algorithm used to explain the model’s decision-making process.

\subsection{Network Architecture}
We design a two-branch network for the hybrid modality of inputs: medical data and 3D body scans. The medical data are represented as a simple vector that can be processed directly by MLP. The 3D body scans, originally stored as polygon meshes, are represented as a set of 2D projections. Twelve views are uniformly sampled around the vertical axis to capture full rotational symmetry, following common practice in multi-view-based methods \cite{su2015multi, chen2021mvt, wang2022ovpt}. Inspired by recent works in multi-view learning, we develop a network branch that integrates low-level CNN with global Transformer to extract features from body shapes \cite{chen2021mvt, wang2022ovpt}. To enhance information exchange between the two branches, we design two separate fusion points in two different ways.

\begin{figure*}[htbp]
\centering
\includegraphics[width=\textwidth]{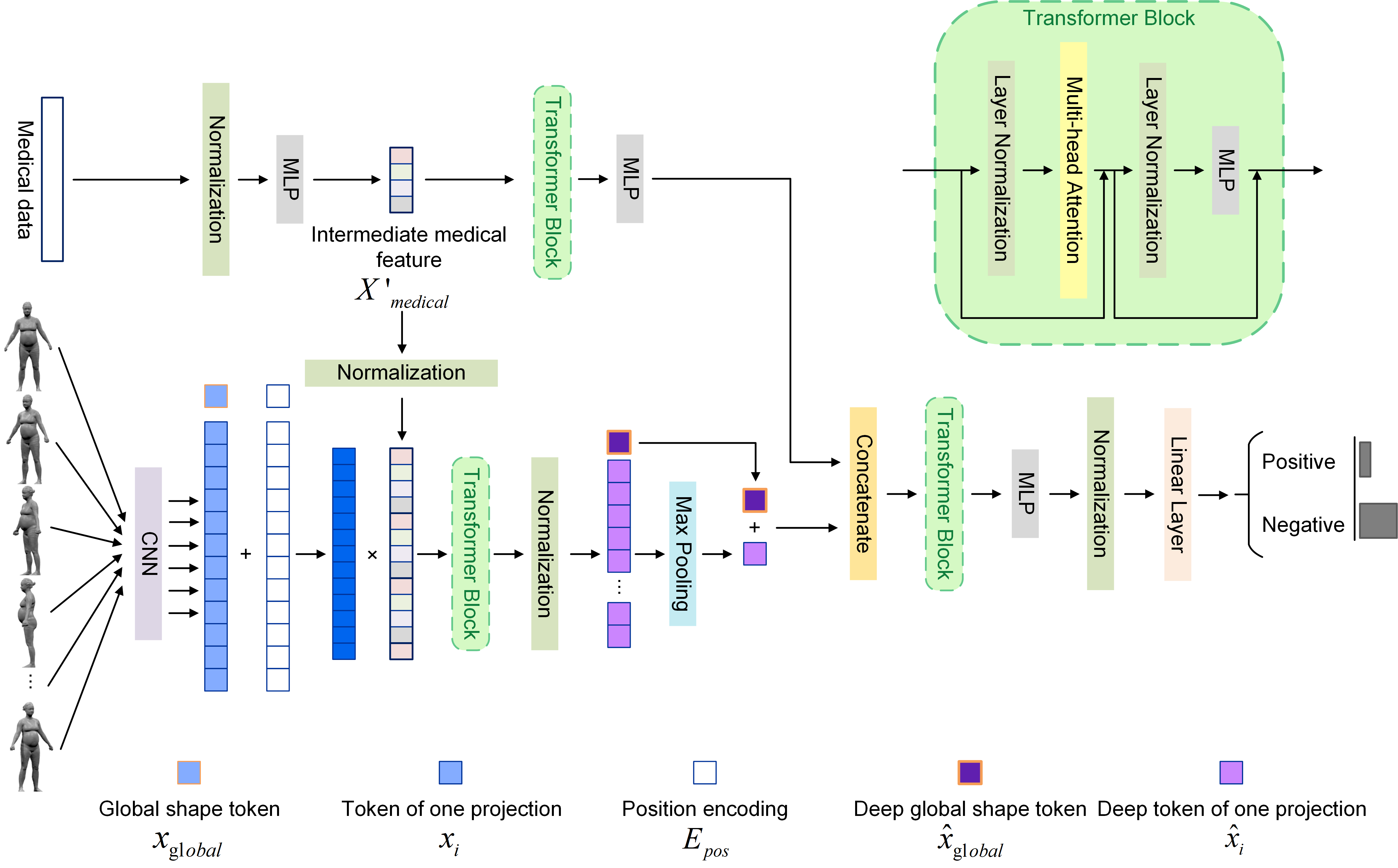}
\caption{Architecture of the proposed hybrid multi-view neural network. The upper stream illustrates the processing of numeric medical features, while the lower stream represents the processing of 3D body features with two-stage fusions of intermediate medical features. The detailed structure of the Transformer Block is shown in the upper-right box, and key vectors are labeled with distinct colors at the bottom for clarity.}
\label{fig:architecture}
\end{figure*}

\textbf{Deep medical feature extraction} 
Given a 1D medical data vector with $d$ variables $X_{medical}\in \mathbb{R}^{1\times d}$,  we normalize the data to the range $[-1,1]$ and then process it using a MLP. Since the obtained output $X'_{medical}$ is also used in the body shape processing branch, as depicted in Fig.~\ref{fig:architecture}, the output dimension of the MLP $d'$ should be chosen properly.

We then apply a Transformer block, which is widely used for its ability to discover deep relationships between features \cite{vaswani2017attention, zheng2024d3bt, li2024infrared}. The Transformer block follows a classical four-layer structure with residual connections, consisting of: (1) layer normalization, (2) multi-head attention layer, (3) another layer normalization, (4) MLP. For the most crucial multi-head attention layer, the input is expanded into three vectors, Query $q$, Key $k$ and Value $v$, by using linear transformation. Attention mechanism is applied by performing a weighted summation of the Value vector, where the weights, also known as an attention map, are computed based on the compatibility between each Query and its corresponding Key \cite{vaswani2017attention}. Suppose we use $N_h$ attention heads, then we take the $j$th attention head $j\in \bigl\{0,1,...,N_h\bigr\}$ as example and calculate:
\begin{equation}
q_j=Linear(LayerInput),q_j\in \mathbb{R}^{1\times \frac{d'}{N_h}},
\end{equation}
\begin{equation}
k_j=Linear(LayerInput),k_j\in \mathbb{R}^{1\times \frac{d'}{N_h}},
\end{equation}
\begin{equation}
v_j=Linear(LayerInput),v_j\in \mathbb{R}^{1\times \frac{d'}{N_h}}.
\end{equation}
The scaled dot-product attention for each head is given by:
\begin{equation}
Output_{ATT_j}=Softmax(\frac{q_jk_j^T}{\sqrt{D_{k_j}}})v_j.
\end{equation}
Then the outputs of all attention heads are concatenated and processed by linear layer to get the output of the multi-head attention layer $Output_{ATT}$:
\begin{equation}
\begin{aligned}
Output_{ATT}=Linear(Concat(Output_{ATT_0},...,\\
Output_{ATT_j},...,Output_{ATT_{N_h}})),\\
j\in \bigl\{0,1,...,N_h\bigr\},Output_{ATT}\in \mathbb{R}^{1\times d'}.
\end{aligned}
\end{equation}
After the processing of the Transformer block and an outer MLP, we obtain the medical feature vector $F_M$ for the next step information fusion and further CS prediction.

\textbf{3D body shape feature extraction} 
To extract features from the projected 3D body scans, we first use pretrained CNN to learn low-level visual information from the 12 projections and get a series of embedded tokens $X_{shape}=\big\{x_0,...,x_i,...,x_{11}\bigr\},x_i\in \mathbb{R}^{1\times C'}$. Here, $C'$ denotes the channel dimension of the embedded tokens, which differs from the original channel dimension $C$ of the $H \times W \times C$ projections. Then we introduce a learnable global shape token $x_{global} \in \mathbb{R}^{1 \times C'}$ and concatenate it with the projection tokens to concentrate the body shape information during learning. We then encode the position of each token to obtain $E_{pos}$ and store the position information by element-wise addition, denoted by the symbol "$\oplus$."
\begin{equation}
\begin{aligned}
X'_{shape}=[x_{global};x_0,...,x_i,...,x_{11}]\oplus E_{pos},\\
X'_{shape}\in \mathbb{R}^{(1+12)\times C'}
\end{aligned}
\end{equation}
After the CNN processing and encoding, we get the pure 3D body shape descriptor, $X'_{shape}$. To enhance the network's understanding of the relationship between body shape and medical features, we introduce the intermediate medical features $X'_{medical}$ as a guide to the body shape processing branch. We employ a soft weight method to combine the intermediate medical features and the shape features. In this approach, the normalized $X'_{medical}$ is replicated to match the number of shape tokens $X'_{shape}$ and fused with the shape tokens through element-wise multiplication \cite{cheng2023ffa}.
\begin{equation}
\begin{aligned}
X''_{shape}=X'_{shape}\otimes Concat(X'_{medical},...,X'_{medical}),\\
X''_{shape}\in \mathbb{R}^{(1+12)\times C'}
\end{aligned}
\end{equation}
Since the dimension of $X'_{medical}$ is much lower than the dimension of $X'_{shape}$, the information $X'_{medical}$ carried is limited, and the shape information is still dominant in the fused vector. After passing the fused result $X''_{shape}$ through the Transformer block and normalization layer, the deep global shape token $\hat{x_{global}}$ aggregates body shape and medical features. However, local features of the projection tokens $\big\{\hat{x_0},...,\hat{x_i},...,\hat{x_{11}}\bigr\}$ may be overlooked when we use single global shape token. Thus, we use the strategy that combines the max-pooling results of all projection tokens and the global shape token to get a more comprehensive shape descriptor $F_S$ \cite{wang2022ovpt}.
\begin{equation}
F_S=max(\hat{x_0},...,\hat{x_i},...,\hat{x_{11}})+\hat{x_{global}},\hat{x_{global}}\in \mathbb{R}^{1\times C'}
\end{equation}

\textbf{Feature fusion and CS prediction}
Although initially represented in different modalities, the 3D body scan has now been transformed into a shape feature vector that aligns in format with the medical feature vector. These two vectors are then concatenated and passed through a final Transformer block followed by an MLP to effectively fuse the deep features.
\begin{equation}
\begin{aligned}
F_{fused}=MLP(Transformer(Concat(F_S,F_M))),\\
F_{fused}\in \mathbb{R}^{1\times (C'+d'')}
\end{aligned}
\end{equation}
A layer normalization and a final linear layer are applied after these operations to make prediction of CS.

\subsection{Soft Margin Triplet-Center Loss for Metric Learning}

Delivery type prediction is a binary classification task, for which cross-entropy loss is commonly used. However, the cross-entropy loss only measures the difference between the ground truth and the predicted value. In contrast,  metric learning can leverage pairwise relationships between samples in the high-dimensional feature space, offering greater efficiency, especially for small datasets and complex biomedical data \cite{liu2024protein}. In this work, we adopt the Soft Margin Triplet-Center Loss (SMTCL) to cluster body shape from the same class closer and keep longer distance between samples from different class \cite{cheng2022soft}. 
\begin{equation}
Loss_{smtcl}=\frac{1}{M}\sum_{i=1}^M{\omega_i(D(f_i,c_{y^i})-D(f_i,c_j))},j\neq y^i
\label{eq:smtcl}
\end{equation}
Here we simplify the original SMTCL for our binary classification task. In \eqref{eq:smtcl}, $M$ denotes the number of samples within a batch, $f_i$ represents the feature of the $i$th sample, and $c_{y^i}$ refers to the center of the class to which the $i$-th sample belongs. $D(f_i,c_{y^i})$ measures the Euclidean distance between a sample and its corresponding class center in the feature space, while $D(f_i,c_j)$ measures the distance between the sample and the center of the opposite class. The parameter $\omega_i$ is a weight dynamically adjusted based on the difference between these two distances: when a sample is closer to the opposite class center and farther from its own class center, it is assigned a higher weight and thus contributes more to the loss. This adaptive weighting mechanism not only avoids tricky hyperparameter tuning but also fully leverages the underlying discrepancies and similarities across individuals. To ensure stable convergence during training, we combine the SMTCL and the standard cross-entropy loss as the final loss function.

\subsection{Integrated Gradients (IG) for Model Explanation}

IG is a feature attribution method designed to explain how neural networks make decisions \cite{sundararajan2017axiomatic}. It addresses the limitations of local gradient-based methods by quantifying feature importance through the integration of gradients along a straight path from a predefined baseline $x'\in \mathbb{R}^n$ to the actual input $x\in \mathbb{R}^n$. In our network, we use black images and zero vectors as baselines for image and medical data inputs respectively, as it represents absence of signal, similar to \cite{sundararajan2017axiomatic}. Denoting our network by a function $F(x)$, the gradient of $F(x)$ with respect to the $i$th input dimension can be expressed as $(\partial F(x))/(\partial x_i)$, and the integrated gradient can be calculated as:
\begin{equation}
IG_i(x)=(x_i-x'_i)\times \int_{\alpha=0}^1{\frac{\partial F(x'+\alpha(x-x'))}{\partial x_i}d\alpha}.
\label{eq:IG}
\end{equation}
In \eqref{eq:IG}, $\alpha \in [0,1]$ is the scaling coefficient used to interpolate between a baseline input $x'$ and the actual input $x$. For image input, the variable $x_i$ represents a pixel; and for medical data input, $x_i$ represents a medical parameter. In the following section, we will visualize the IG results in different ways for the different input modalities.

\section{Experiment}
\label{sec:experiment}

\subsection{Classification performance}
\begin{table}[bt]
\caption{Cross-validation performance of different methods for CS prediction}
\centering
\begin{tabular}{l*{5}{p{11mm}}p{14mm}}
\toprule
Method & accuracy & precision & recall & specificity & F1 & AUC (p-value) \\
\midrule
LR (MD)                                                & 74.67\% & 52.00\%   & 65.00\% & 78.18\%     & 0.578   & 0.658(0.037) \\

LR (MD \& PCs)                                 & 70.67\% & 45.83\%   & 55.00\% & 76.36\%     & 0.500   & 0.702(0.008) \\

RF (MD \& PCs)                                  & 69.33\% & 44.44\%   & 60.00\% & 72.73\%     & 0.511   & 0.690(0.012) \\

SVC (MD \& PCs)                                 & 66.67\% & 42.42\%   & \textbf{70.00\%} & 65.45\%     & 0.528   & 0.707(0.006) \\

BPNN (MD \& PCs)                                & 70.67\% & 45.83\%   & 55.00\% & 76.36\%     & 0.500   & 0.707(0.006) \\

Cheng et al. \cite{cheng2025maternal} (MD \& LCs) & 72.00\% & 48.28\%   & \textbf{70.00\%} & 72.73\%     & 0.571   & 0.691(0.012)  \\

MVCNN \cite{su2015multi} (MD \& 3D scan)        & 74.67\% & 53.33\% & 40.00\% & 87.27\% & 0.457 & 0.706(0.007) \\

OVPT \cite{wang2022ovpt} (MD \& 3D scan)        & 77.33\% & 61.54\%   & 40.00\% & 90.91\%     & 0.485   & 0.664(0.031)  \\

PCT \cite{guo2021pct} (MD \& 3D scan)                             & 74.67\% & 52.63\%   & 50.00\% & 83.64\%     & 0.513   & 0.701(0.008)  \\

MvBody (MD \& 3D scan)                                  & \textbf{82.67\%} & \textbf{73.33\%}   & 55.00\% & \textbf{92.73\%}     & \textbf{0.629}   & \textbf{0.746(0.001)}  \\
\bottomrule
\multicolumn{7}{l}{Note: Model inputs are listed in parentheses following the model names; MD denotes "medical data";}\\
\multicolumn{7}{l}{\hspace{0.8cm} PCs refer the first four PCs of anthropometric measurements; LCs refer abdominal level}\\
\multicolumn{7}{l}{\hspace{0.8cm} circumferences extracted from 3D human body;}\\
\end{tabular}
\label{tab:classification}
\end{table}

\begin{table}[bt]
\caption{Performance of top methods for CS prediction on the test set}
\centering
\begin{tabular}{l*{5}{p{11mm}}p{14mm}}
\toprule
Method & accuracy & precision & recall & specificity & F1 & AUC (p-value) \\
\midrule
SVC (MD \& PCs)                                & 76.92\% & 40.00\%   & 40.00\% & 85.71\%     & 0.400   & 0.705(0.161) \\

BPNN (MD \& PCs)                                & \textbf{84.62\%} & \textbf{66.67\%}   & 40.00\%  &  \textbf{95.24\%}     & 0.500   & 0.657(0.284) \\

MvBody (MD \& 3D scan)                                  & \textbf{84.62\%} & 60.00\%   &  \textbf{60.00\%}  &  90.48\%     & \textbf{0.600}   & \textbf{0.724(0.126)}  \\

\bottomrule
\multicolumn{7}{l}{Note: Model inputs are listed in parentheses following the model names; MD denotes "medical data";}\\
\multicolumn{7}{l}{\hspace{0.8cm} PCs refer the first four PCs of anthropometric measurements.}\\
\end{tabular}
\label{tab:classification_test}
\end{table}

\begin{figure*}[htbp]
\centering
\includegraphics[width=\textwidth]{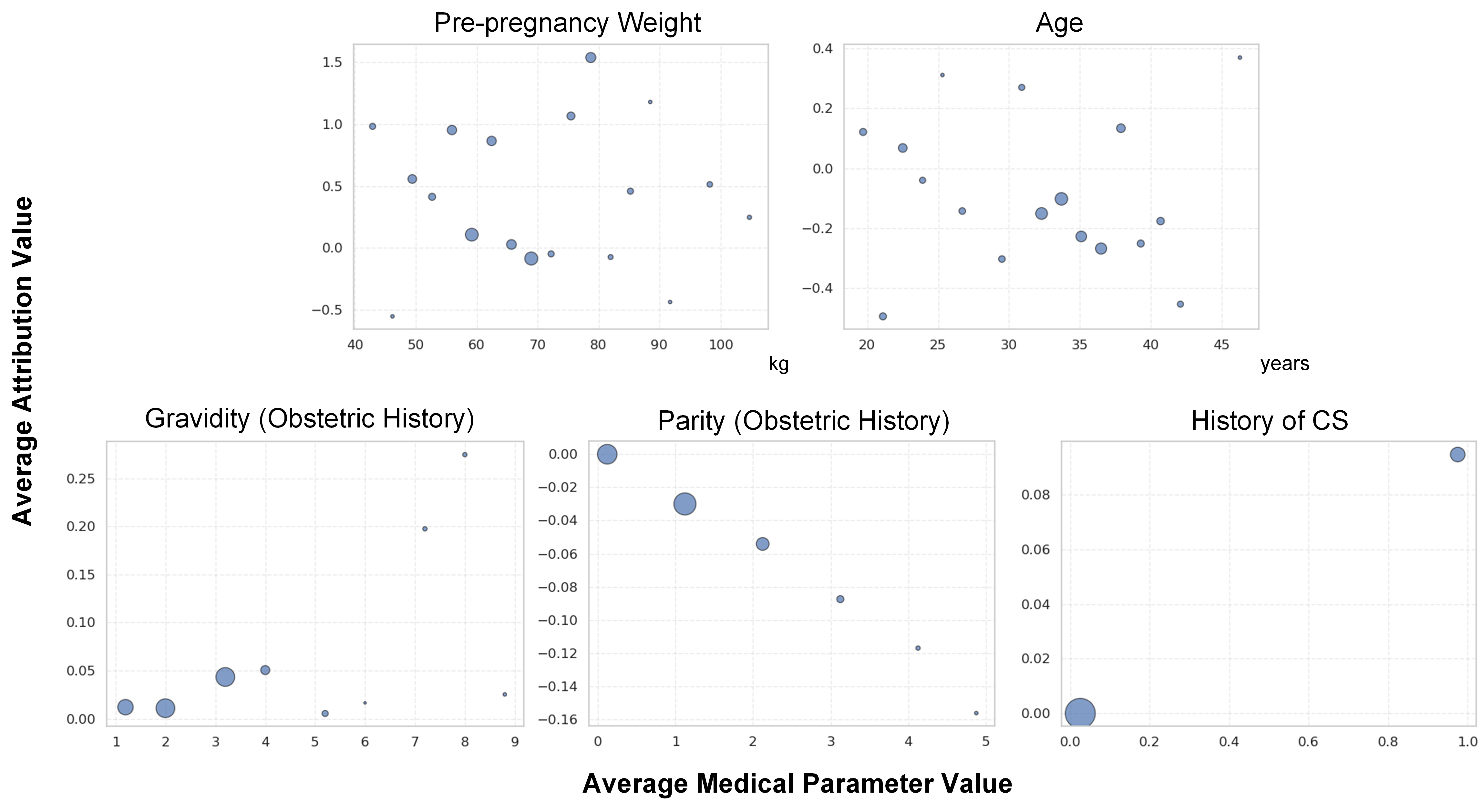}
\caption{Distributions of the most significant medical parameters' attributions.}
\label{fig:bubble}
\end{figure*}

To evaluate our MvBody for CS prediction, we used accuracy, precision, recall, specificity, F1 score and the Area Under the Curve of a Receiver Operating Characteristic curve (AUC-ROC) as evaluation metrics, and computed the p-value for AUC-ROC using the Mann-Whitney U test. Popular machine learning models including LR, RF, Support Vector Classification (SVC) and Backpropagation Neural Network (BPNN) were evaluated for comparison. For these algorithms, we performed principal component analysis on the anthropometric measurements generated by Fit3D and used the first four principal components (PCs) along with the medical data as input. We also used an LR model with only medical data as input as the baseline method. Advanced 3D analysis models were also adapted to our dataset for comparison, including MVCNN \cite{su2015multi}, OVPT \cite{wang2022ovpt}, and PCT \cite{guo2021pct}. Among them, the first two are multi-view-based models, while the latter two incorporate Transformer architectures. In addition, we tested the latest 3D body shape-based predictive model \cite{cheng2025maternal} for comparison, which uses abdominal level circumferences combined with the same medical data as input. 

In our implementation, the CNN block was VGG11 \cite{simonyan2014very}, which accounted for 94\% of the network’s parameters and was pretrained on the large-scaled ImageNet \cite{deng2009imagenet}. After projection, a total of 2832 images from 236 scans across 101 participants were used in our experiments. Of these participants, 75\% were randomly selected for 5-fold cross-validation, while the remaining 25\% were reserved as an independent test set. For participants with multiple scans, the highest prediction score among their scans was used as the final CS risk score during evaluation. 

Tab.~\ref{tab:classification} summarizes the performances of different methods in predicting CS on 5-fold cross-validation dataset. From the table, we observed that the basic LR model using only self-reported medical data achieved relatively good performance in CS risk prediction. However, incorporating anthropometric measurements did not significantly improve predictive performance. Even when employing more advanced machine learning algorithms, we could only see higher recall rate of 70.00\% yielded by SVC and the latest shape-based method. This might be attributed to the anthropometric measurements losing detailed shape information, and the limited capabilities of traditional machine learning algorithms in extracting useful body shape features for accurate CS prediction. Although incorporating advanced network architectures, the three 3D analysis models still exhibited limited performance, which may be attributed to insufficient fusion between medical and 3D shape features, as well as the lack of training strategies tailored to small and imbalanced datasets. Notably, our MvBody achieved the highest scores in 5 out of 6 metrics, with the exception of the recall of 55.00\%. But for all methods, it was hard to identify CS caused by malpresentation. We further evaluated the top three methods with the highest AUC-ROC scores on the independent test set, training them on the full cross-validation dataset. As shown in Tab.~\ref{tab:classification_test}, the results are consistent between cross-validation and the independent test set that our method has the best performance with AUC-ROC drops slightly.

\subsection{Visualization of Model Explanations}
We used IG to analyze the contribution of each medical parameter and body projection separately in the group for cross-validation. To explore how the medical parameters influence the prediction of CS, we generated a series of bubble charts for the most significant parameters based on the magnitude of their attributions, as shown in Fig.~\ref{fig:bubble}. For numeric data, samples were grouped into clusters according to medical parameter values, while for binary categorical data, they were naturally divided into 2 clusters. The size of each bubble indicates the number of samples in the cluster. In each bubble chart, the x-axis represents the medical parameter value, and y-axis represents its attribution value. Positive attributions indicate contributions to the positive class (CS delivery), and negative attributions to the negative class (vaginal delivery). The magnitude of the attribution reflects the strength of each feature’s influence.

\begin{figure}[]
\centering
\includegraphics[width=\textwidth]{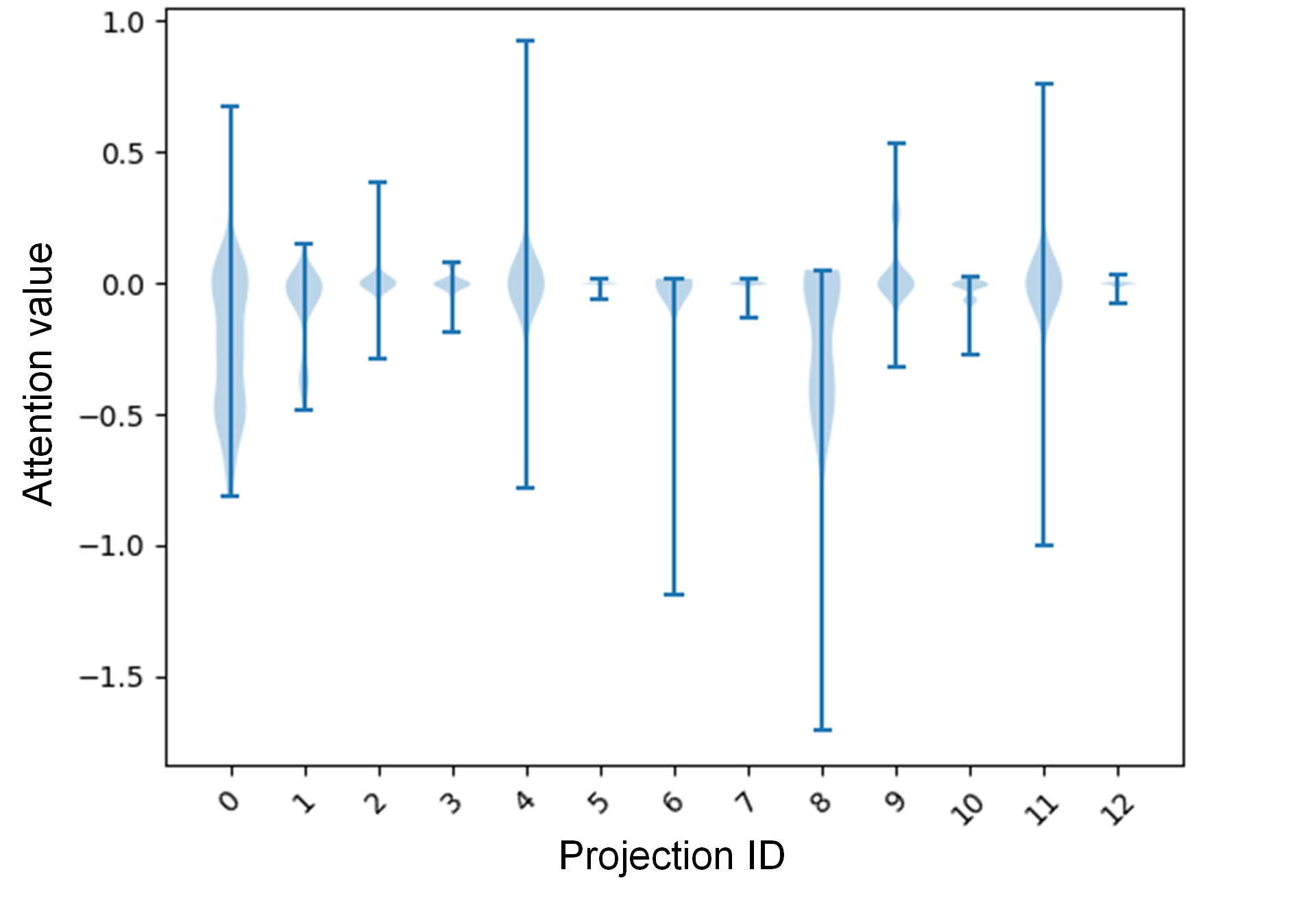}
\caption{Distribution of different projection tokens' attributions.}
\label{fig:violin}
\end{figure}

The medical parameters identified as most significant included “pre-pregnancy weight”, “age”, “obstetric history-gravidity”, “obstetric history-parity” and “history of CS”, each showing a distinctly broader range of attribution values. The charts revealed that a higher number of gravidity (pregnancies) increased the likelihood of a positive prediction, while a higher number of parity (deliveries) correlated with a lower CS risk. The impact of CS history is also consistent with clinical expectations as a history of CS increases the likelihood of a repeat CS in subsequent pregnancies. However, the effects of pre-pregnancy weight and age exhibit more complex patterns. These factors may interact with other variables, influencing their correlation with CS risk. We can only roughly conclude that individuals with a pre-pregnancy weight around 70 kg were predicted to have a lower risk of CS, while pregnancies at maternal ages below 20 or above 45 appeared more likely to require a CS. Furthermore, we observed that having chronic asthma or a history of gestational hypertension slightly increased the probability of CS.

\begin{figure*}[htbp]
\centering
\includegraphics[width=\textwidth]{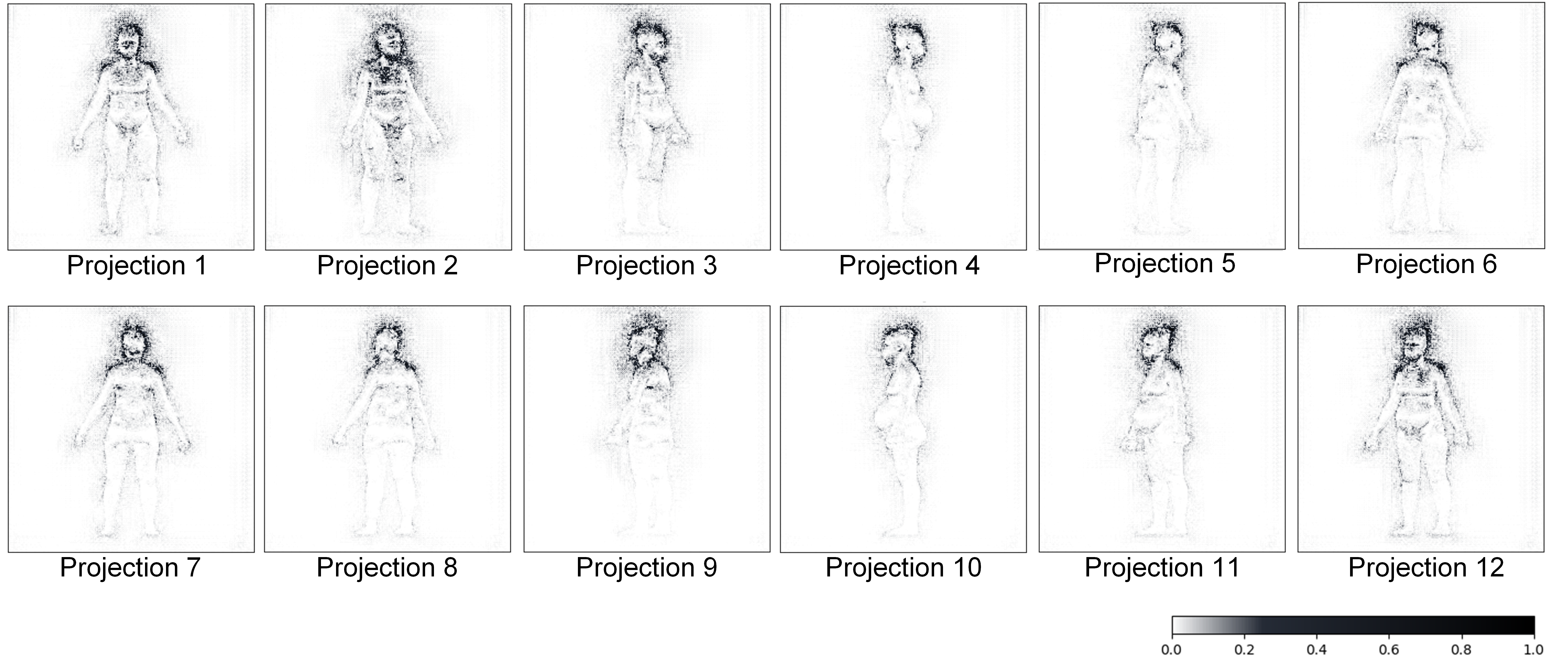}
\caption{Pixel-level attributions of input projections. Grayscale intensity represents the magnitude of attribution at each pixel.}
\label{fig:projection}
\end{figure*}

Regarding the body projections, we analyzed their corresponding tokens alongside the global shape token using a violin plot (Fig.~\ref{fig:violin}), which directly illustrated the attributions. We assigned ID 0 to the global shape token and IDs 1 through 12 to the projection tokens according to rotation angle sequence. The vertical bar in the violin chart represents the range of attribution values of all samples for each projection token. And the width of the shaded area indicates the density of the samples at each attribution value. Similarly, a larger magnitude and more even distribution of attribution values indicate that a token carries more information for body shape representation and CS prediction.

From Fig.~\ref{fig:violin} we observed that the global shape token aggregated crucial information from all projections. In addition, projections 1, 2, 4, 6, 8, 9 and 11 also appeared to capture significant features. To further investigate the viewing angles of these projections and identify which specific regions within them contributed more to the prediction, we visualized the pixel-level attribution distribution of the input 2D projections in Fig.~\ref{fig:projection}.

By aligning the IDs of the tokens and IDs of the visualized attributions in 2D projections in Fig.~\ref{fig:violin} and Fig.~\ref{fig:projection}, and considering the bilateral symmetry of the human body, we observed that the “important” projections covered all views. Upon inspecting the grayscale attribution map, we found that pixels near the body's edges play important roles. These pixels represented the potential body silhouette, from which useful anthropometric measurements could be derived for the subsequent prediction. Another interesting finding was that regions around the shoulders and head showed stronger attributions than the abdominal region, contrary to our initial expectation but consistent with observations reported in previous studies \cite{solomon2018age,alijahan2014diagnostic,liselele2000maternal}. Thus we designed an ablation experiment by removing these regions from the input to explore whether the shapes of head and shoulders introduce useful information. Details of this experiment are discussed in the following subsection.

\subsection{Ablation Study}
\begin{table}[htbp]
\caption{Performance of MvBody using projections with and without specific body parts}
  \centering
\begin{tabular}{l*{6}c}
\toprule
Input & accuracy & precision & recall & specificity & F1 & AUC \\
\midrule
3D scan w/o legs             & 78.67\%	& 62.50\%	& 50.00\% & 89.09\%	   & 0.556	& 0.695	\\

3D scan w/o head \& shoulders             & 70.67\%	& 43.75\%	& 35.00\% & 83.64\%	   & 0.389	& 0.594	\\

3D scan w/o head                         & 70.67\% & 45.45\%   & 50.00\% & 78.18\%     & 0.476   & 0.644 \\

3D scan                                  & \textbf{82.67\%} & \textbf{73.33\%}   & \textbf{55.00\%} & \textbf{92.73\%}     & \textbf{0.629}   & \textbf{0.746}  \\
\bottomrule
\end{tabular}
  \label{tab:ablation_head}
\end{table}

\begin{table}[htbp]

\caption{Performance of our method with different component combinations }
  \centering
\begin{tabular}{*{9}c}
\toprule
\multicolumn{3}{c}{Components}& accuracy & precision & recall & specificity & F1 & AUC \\ 
SMTCL & Transformer blocks & First stage fusion & & & & & & \\
\midrule
 &   &   & 74.67\% & 53.33\% & 40.00\% & 87.27\% & 0.457 & 0.706 \\

$\surd$ & & & 74.67\% & 52.94\% & 45.00\% & 85.45\% & 0.487 & 0.640 \\

  & $\surd$ & & 80.00\% & 69.23\% & 45.00\% & \textbf{92.73\%} & 0.545 & 0.686 \\

  & & $\surd$ & 80.00\% & 61.90\% & \textbf{65.00\%} & 85.45\% & \textbf{0.634} & 0.745 \\

$\surd$ & $\surd$ & $\surd$ & \textbf{82.67\%} & \textbf{73.33\%} & 55.00\% & \textbf{92.73\%} & 0.629 & \textbf{0.746} \\
\bottomrule
\end{tabular}
    \label{tab:ablation_components}
\end{table}

Based on the visualization results in Fig.~\ref{fig:projection}, we investigated whether the head region and shoulder region brought noise or provides useful information. We also evaluated the impact on plain algorithm when replacing regular cross-entropy loss with the SMTCL, when adding the Transformer blocks, and when using the first-stage fusion. We used the same evaluation metrics and tested the algorithms on the same prediction task in the group for cross-validation. 

The comparison results when using body shape projections with specific region covered are shown in Tab.~\ref{tab:ablation_head}. Compared with using full body projections as input, when using projections without head and shoulders, the prediction performance dropped significantly across all metrics. When we only removed the head and kept the shoulders, the performance increased except for the specificity. In contrast, using body shape with legs covered did not cause a significant decrease in performance. These results indicated that head and shoulders shapes contain important information for CS prediction. 

The performances of different algorithm configurations with the same inputs are presented in Tab.~\ref{tab:ablation_components}. We concluded that using SMTCL instead of regular cross-entropy loss improved the identification of positive samples. The result aligned with the function of SMTCL, which emphasized correctly classifying the “hard samples”. Significant improvement was seen across most metrics when Transformer blocks were added, highlighting their strong ability to explore and integrate useful contextual information. Comparing the third and sixth rows, we observed a noticeable improvement in performance, especially in recall, which demonstrated the importance of incorporating medical features into body shape learning at an early stage to better distinguish high-risk population. Overall, SMTCL, Transformer blocks and the first-stage fusion significantly contribute to the performance of our model, with their combination yielding the best results.

\section{Conclusion}
\label{sec:conclusion}

In this study, we introduced MvBody, a novel multi-view Transformer-based neural network designed to predict CS risk using self-reported medical data and 3D optical body scans. Pretraining and metric learning strategies were used to address the challenge of small sample size and high data dimensionality. Compared to other well-performed methods, the MvBody demonstrated superior performance, achieving an accuracy of 84.62\%, recall of 60.00\%, specificity of 90.48\%, and AUC-ROC of 0.724. We further explained the results using the IG algorithm and visualized the attributions separately for medical parameters and 2D projections. Our analysis revealed that pre-pregnancy weight, age, gravidity, parity, CS history, and the shape of the maternal head and shoulders play important roles in making the accurate prediction.

Despite its promising results, this study has several limitations. First, the dataset size is relatively small. Although targeted training strategies were applied to mitigate this issue, a larger dataset could enhance the robustness and accuracy of predictions across diverse populations. Second, this study serves as a pilot investigation toward more affordable future applications. It currently relies on scans obtained from commercial 3D scanners. Future work will explore the feasibility of using smartphone-based 3D scanning solutions and adapt the model accordingly to support broader applications.

\section*{Acknowledgment}
Research reported in this publication was supported by the National Institute of Diabetes And Digestive and Kidney Diseases under Award Number R01DK129809 and by the National Institute on Aging under Award Number R56AG089080 of the National Institutes of Health. The content is solely the responsibility of the authors and does not necessarily represent the official views of the National Institutes of Health.

\end{document}